\documentclass[10pt,twocolumn]{article}

\usepackage[letterpaper,margin=0.72in]{geometry}
\usepackage{amsmath,amssymb}
\usepackage{booktabs}
\usepackage{graphicx}
\usepackage{microtype}
\usepackage{multirow}
\usepackage{natbib}
\usepackage{xcolor}
\usepackage[hidelinks]{hyperref}
\usepackage[font=small,labelfont=bf]{caption}
\usepackage{subcaption}
\usepackage{enumitem}

\graphicspath{{figures/}}
\setlist{nosep,leftmargin=*}

\makeatletter
\renewcommand{\@maketitle}{%
  \newpage
  \null
  \vskip 0.5em%
  \begin{center}%
    {\LARGE \@title \par}%
    \vskip 0.8em%
    {\large \lineskip 0.25em\begin{tabular}[t]{c}\@author\end{tabular}\par}%
  \end{center}%
  \par
  \vskip 0.7em}
\makeatother

\title{Evidence-Order Calibration for Selective Visual Reasoning\\
under Progressive Loss of Question-Critical Evidence}
\author{MUHAMATHU AMEER ALI AACAAS MUHAMATH\\
Department of Electrical Engineering\\
University of Moratuwa\\
Moratuwa, Sri Lanka\\
\texttt{muhamathmaaa.23@uom.lk}}
\date{}

\begin{document}
\maketitle

\begin{abstract}
Vision--language model (VLM) confidence may change in aggregate when visual evidence is degraded while remaining structurally inconsistent within individual examples. We study answer-level reliability along five-step, question-conditioned evidence-loss trajectories. Using a frozen Qwen2.5-VL-3B-Instruct model, we construct 176 accepted GQA-derived trajectories (880 masking conditions) by progressively masking scene-graph-localized question-critical regions. Native sequence confidence has an evidence monotonicity violation rate (EMVR) of 0.436, and 92.0\% of trajectories contain at least one adjacent violation. A matched non-critical-region control shows that full critical masking reduces accuracy by 28.2 percentage points, compared with 0.6 points for equally sized non-critical masks; the paired difference is 27.6 points (95\% CI [20.0, 34.7]). We train a lightweight post-hoc reliability head on frozen hidden states, sequence confidence, and entropy. Adding evidence-order supervision to binary cross-entropy (BCE) reduces masking EMVR from 0.330 to 0.303 (paired difference $-0.027$, 95\% CI [$-0.044$, $-0.010$]). The same mask-trained objective reduces EMVR from 0.449 to 0.402 on held-out question IDs under unseen local Gaussian blur (difference $-0.0468$, 95\% CI [$-0.0739$, $-0.0199$]). AUROC, Brier, and AURC differences between the two learned heads are statistically inconclusive, and native confidence remains stronger for selective-risk ranking. The results separate evidence-order consistency from conventional correctness discrimination rather than establishing generic confidence superiority.
\end{abstract}

\section{Introduction}
\label{sec:introduction}

Vision--language models can answer visual questions fluently even when the pixels needed to justify an answer are weak, absent, or contradicted by language priors. Existing work measures object hallucination, visual dependence, uncertainty, and abstention in complementary ways \citep{li2023pope,leng2023vcd,favero2024m3id,srinivasan2024recoverr}. We ask a narrower behavioral question: \emph{when progressively more of the visual evidence required by this particular question is removed, does answer reliability respect the resulting order?}

For a clean image--question pair and a question-critical region, we create an ordered sequence of local evidence states at severities $\lambda\in\{0,.25,.5,.75,1\}$. A reliability score consistent with the intervention should satisfy
\begin{equation}
R_0 \ge R_{.25} \ge R_{.5} \ge R_{.75} \ge R_1.
\label{eq:desired-order}
\end{equation}
This requirement differs from ordinary correctness discrimination. A score can rank correct answers above incorrect answers on average and still increase as evidence for one question is removed. Conversely, a highly order-consistent score need not be the best global selector of correct answers. Our experiments show both sides of this distinction.

\begin{figure*}[t]
    \centering
    \includegraphics[width=\textwidth]{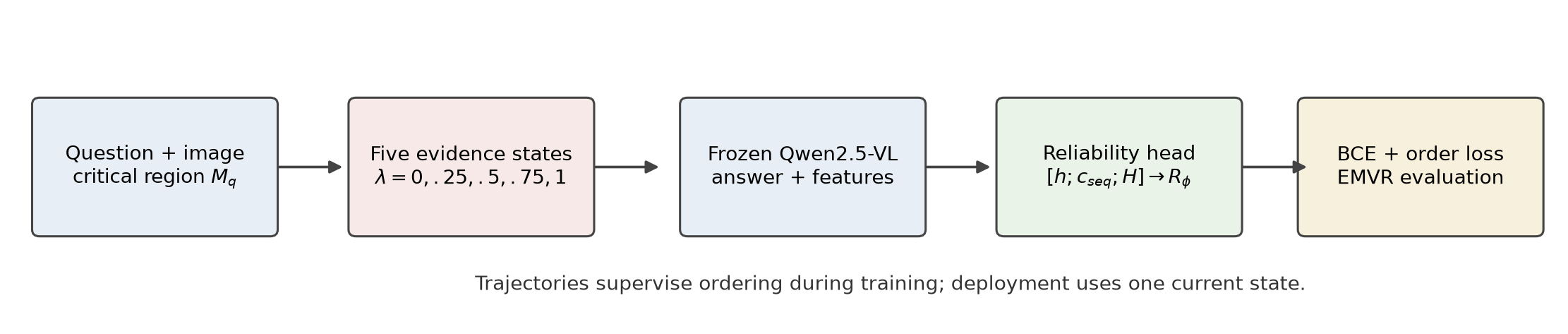}
    \caption{Study design. Progressive, local, question-critical evidence loss produces supervision for a post-hoc reliability head while the answer-generating VLM remains frozen.}
    \label{fig:pipeline}
\end{figure*}

We use GQA scene graphs \citep{hudson2019gqa,krishna2016visualgenome} to associate questions with object boxes and construct five-level local masking trajectories. The answer model, Qwen2.5-VL-3B-Instruct \citep{bai2025qwen25vl}, remains frozen. We extract its answer, native sequence log-confidence, predictive entropy, and final-layer hidden representation. A linear post-hoc head predicts strict normalized answer correctness; an evidence-order hinge loss additionally supervises the relative logits between adjacent severity states within each trajectory. Figure~\ref{fig:pipeline} summarizes the protocol. At deployment, the head consumes only the features of the current image, question, and generated answer; the complete trajectory is a source of training and evaluation supervision, not a deployment requirement.

The evidence manipulation is validated with an equally sized matched non-critical-region control. Destroying critical evidence causes approximately 28 percentage points of accuracy degradation, whereas full non-critical masking causes almost none. The primary learned comparison then isolates the training objective: BCE plus evidence-order supervision significantly lowers the evidence monotonicity violation rate (EMVR) relative to a BCE-only head. Most importantly, this structural benefit transfers from masking to an unseen local Gaussian-blur degradation on held-out question IDs. The transfer does not establish universal calibration: conventional AUROC, Brier, and selective-risk differences are limited or statistically inconclusive, and native confidence retains the best AURC.

Our contributions are threefold:
\begin{enumerate}
    \item \textbf{Controlled evidence trajectory protocol.} We construct progressive, question-critical local evidence-loss trajectories and a task-specific diagnostic of order violations, together with matched non-critical controls.
    \item \textbf{Evidence-order reliability estimator.} We train a lightweight post-hoc estimator with correctness and ordinal evidence-order supervision while keeping the VLM frozen.
    \item \textbf{Cross-degradation empirical analysis.} Evidence-order supervision reduces violations under masking and retains a statistically supported reduction on unseen local blur, while conventional discrimination and selective-risk gains remain limited.
\end{enumerate}

We do not claim the first use of ordinal VLM calibration, hidden-state confidence probes, region masking, evidence/non-evidence intervention, or abstention. The contribution is their use in a specific question-conditioned, multi-severity answer-reliability protocol and the accompanying transfer test.

\section{Related Work}
\label{sec:related}

\paragraph{Hallucination and visual dependence.}
POPE established a polling-based object-hallucination evaluation for LVLMs \citep{li2023pope}. Visual Contrastive Decoding (VCD) contrasts outputs from original and distorted visual inputs \citep{leng2023vcd}; OPERA modifies decoding through an over-trust penalty and retrospective allocation \citep{huang2023opera}; and M3ID strengthens visual influence through mutual-information decoding \citep{favero2024m3id}. These methods motivate careful measurement of whether answers depend on images, but our target is neither caption hallucination nor answer-model decoding. We leave generation frozen and estimate answer reliability after generation.

\paragraph{Multimodal uncertainty and selective prediction.}
ReCoVERR studies selective visual reasoning and uses additional evidence gathering to reduce unnecessary abstention \citep{srinivasan2024recoverr}. HARMONY combines hidden activations with model-output information for VLM uncertainty estimation \citep{mushtaq2025harmony}, establishing that such features are useful but not themselves novel here. CSP applies object-level semantic perturbations to verbalized confidence calibration \citep{zhao2025csp}. Recent answerability benchmarks explicitly study insufficient evidence: MM-AQA constructs multimodal unanswerable instances \citep{madhusudhan2026mmaqa}, while TRAPSBench probes epistemic restraint and the hidden-state representation of answerability \citep{pramono2026trapsbench}. Our score can support abstention, but our core diagnostic is ordering along controlled local evidence loss rather than generic answerability.

\paragraph{Visual evidence intervention and causal grounding.}
Contrastive Region Masking diagnoses step-level dependence by masking annotated visual regions \citep{chaturvedi2025crm}. Evidence-RL compares object-centric evidence regions with matched non-evidence regions inside answer-model reinforcement learning \citep{huang2026evidencerl}. These works make region intervention and relevant-versus-noncritical controls clear prior ingredients. In our study, masking is an experimental instrument; the answer model is not post-trained, and the learned component is an answer-level reliability estimator.

\paragraph{Ordinal calibration and confidence ranking.}
VORD calibrates LVLM token predictions using ordinal relations between modified-image pairs and includes a trainable ordinal objective \citep{neo2024vord}. It is therefore incorrect to present ordinal VLM calibration as new. BICR trains a lightweight confidence probe on frozen LVLM representations using a real-image versus globally blinded-image ranking contrast \citep{khanmohammadi2026bicr}. Relative to these closest neighbors, our intervention is local and tied to the evidence required by a specific question; supervision spans five progressive severities rather than a generic modified pair or binary real/blind contrast; evaluation is answer-level; and the central transfer test trains on masks and tests on local blur.

Existing literature thus establishes the ingredients individually: hallucination evaluation, corrupted-image decoding, hidden-state reliability, selective prediction, region intervention, binary blind-image ranking, and ordinal calibration. We investigate whether \emph{progressive question-critical local evidence loss} supplies useful ordinal supervision for a frozen-model reliability head, and whether the learned order survives a change in degradation mechanism.

\section{Method}
\label{sec:method}

\subsection{Question-conditioned evidence trajectories}

Let $f_\theta$ be a frozen VLM, $I$ an image, $q$ a question, and $M_q$ the bounding box of the object judged critical for answering $q$. A corruption operator $C(I,M_q;\lambda)$ removes a severity-dependent fraction of this region. For masking, the side lengths of the centered gray rectangle scale as $\sqrt{\lambda}$, so its area scales linearly with $\lambda$. At $\lambda=0$ the image is unchanged; at $\lambda=1$ the full box is covered. Applying $f_\theta$ at the five severities yields answer and feature tuples $(y_{i,k},x_{i,k})$ for trajectory $i$ and severity index $k$.

The strict normalized correctness label is $z_{i,k}\in\{0,1\}$. We use the frozen VLM's final language-layer representation $h_{i,k}\in\mathbb{R}^{2048}$, mean semantic-token sequence log-confidence $c_{i,k}$, and mean predictive entropy $H_{i,k}$. All tokenizer special/control tokens are excluded from the two token-level aggregates. The combined feature vector is
\begin{equation}
x_{i,k}=[h_{i,k};c_{i,k};H_{i,k}]\in\mathbb{R}^{2050}.
\end{equation}

\subsection{Post-hoc reliability head}

A linear head produces logit $g_{i,k}=w^\top x_{i,k}+b$ and reliability
\begin{equation}
R_\phi(x_{i,k})=\sigma(g_{i,k}).
\end{equation}
The answer-generating VLM is unchanged. The BCE-only baseline minimizes
\begin{equation}
\mathcal{L}_{\mathrm{cls}}=-\sum_{i,k}\left[z_{i,k}\log R_{i,k}+(1-z_{i,k})\log(1-R_{i,k})\right].
\end{equation}
The order-aware model adds an ordinal hinge loss over adjacent
severity pairs within each trajectory:
\begin{equation}
\mathcal{L}_{\mathrm{ord}}
=
\sum_i \sum_{k=0}^{3}
\max\left(0,\,
m-\left[g_{i,k}-g_{i,k+1}\right]
\right).
\end{equation}
The frozen setting is
\begin{equation}
\mathcal{L}=\mathcal{L}_{\mathrm{cls}}+\mu\mathcal{L}_{\mathrm{ord}}+\alpha\lVert w\rVert_2^2,
\end{equation}
with $\mu=0.3$, margin $m=0.05$, L2 setting $1.0$, and seed 42. These values were fixed before the final control and transfer analyses.

\subsection{Evidence monotonicity violation rate}

For adjacent severities, a violation occurs when reliability rises after more evidence is removed:
\begin{equation}
v_{i,k}=\mathbb{1}[R_{i,k+1}>R_{i,k}].
\end{equation}
With five states per trajectory,
\begin{equation}
\mathrm{EMVR}=\frac{1}{4N}\sum_{i=1}^N\sum_{k=0}^{3}v_{i,k}.
\label{eq:emvr}
\end{equation}
Lower is better. EMVR is a task-specific behavioral diagnostic of evidence-order consistency, not a universal uncertainty or calibration metric. We also report the fraction of trajectories with at least one violation, correctness discrimination (AUROC/AUPRC), Brier score, and area under the risk--coverage curve (AURC).

\subsection{Deployment interpretation}

Trajectories are required to construct the order loss and evaluate EMVR. Once trained, the post-hoc head operates on a single current image--question--answer state. It does not mask the image repeatedly or generate a full trajectory at deployment time.

\section{Experimental Setup}
\label{sec:setup}

\paragraph{Data and grounding.}
We derive a controlled subset from GQA balanced validation questions \citep{hudson2019gqa}, whose scene graphs inherit dense object geometry from Visual Genome \citep{krishna2016visualgenome}. Candidate questions focus on color, identity, material, and shape. Each trajectory uses the first mapped critical object box. The initial scaled set contained 250 candidates: automated visual quality control accepted 176, rejected 71, and marked 3 unsure (70.4\% acceptance). A limited ten-example human spot-check supplemented, but did not replace, this automated audit. We do not describe the benchmark as fully manually audited.

The accepted manifest is frozen at \path{data/gqa/manifests/gqa_evidence_scaled_accepted.json}. The main masking experiment contains 176 trajectories and 880 conditions. A subsequent P17 cleanup corrected 205 category labels without changing any correctness labels; category-specific analysis uses the corrected taxonomy. Overall metrics are unaffected.

\paragraph{Model and generation.}
The primary model is the non-quantized \texttt{Qwen/Qwen2.5-VL-3B-Instruct} checkpoint \citep{bai2025qwen25vl}. Generation is deterministic (\texttt{do\_sample=False}) with a short-answer prompt. All main experiments freeze the VLM and train only lightweight linear heads. The executed environment used Python 3.14.5, PyTorch 2.11.0+cu128, Transformers 5.16.1, and an NVIDIA GeForce RTX 5060 Laptop GPU with 8,151 MiB; \texttt{device\_map=auto} permitted CPU/disk offload where necessary.

\paragraph{Matched non-critical control.}
For 170 of 176 accepted samples, a deterministic grid search finds a fully in-bounds box with the same width and height as the critical box and critical-control IoU at most 0.05. Candidates minimize total intersection ratio with non-critical scene-graph objects, then maximize center distance from the critical box. Six samples have no defensible same-size placement and are excluded only from the paired control. The mean critical-control IoU is 0.00506. Thirty-two controls have zero annotated-object overlap; because dense/nested annotations remain, we call them \emph{matched non-critical regions}, not pure background.

\paragraph{Held-out blur transfer.}
The unseen degradation applies Gaussian blur only within $M_q$ with radius
\begin{equation}
r(\lambda)=\min\left(24,0.15\lambda\min(w,h)\right).
\end{equation}
Models are trained only on masking trajectories and evaluated on blur trajectories for held-out question IDs using grouped cross-validation. No blur examples or blur-specific hyperparameter tuning enter training.

\paragraph{Splits and uncertainty.}
Question IDs define groups so all five states of a trajectory remain together. The final comparisons use out-of-fold predictions. Uncertainty for method differences is estimated with 1,000 paired bootstrap resamples (seed 42) at the whole-question-trajectory level; severity rows are never resampled independently. We report percentile 95\% intervals and avoid p-values.

\section{Results}
\label{sec:results}

\subsection{Critical evidence loss degrades task performance}

\begin{figure*}[t]
    \centering
    \includegraphics[width=\textwidth]{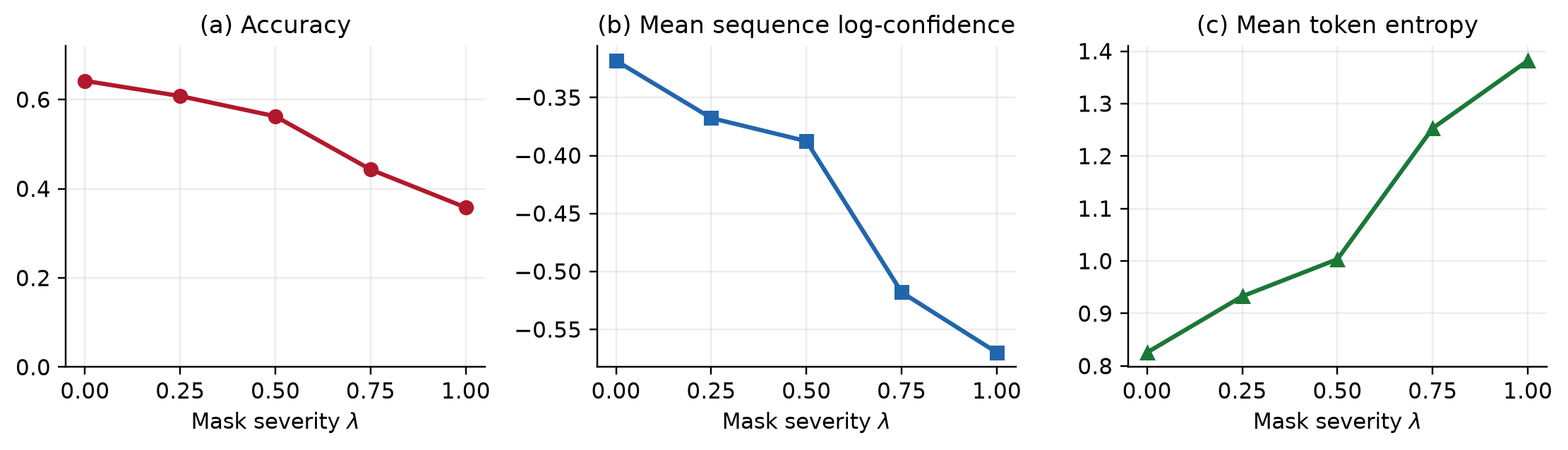}
    \caption{Aggregate behavior under progressive masking of question-critical evidence. Accuracy and sequence log-confidence decrease while entropy rises.}
    \label{fig:severity}
\end{figure*}

Figure~\ref{fig:severity} shows the aggregate phenomenon. Accuracy falls from 0.642 on clean images to 0.358 under full critical masking, a 28.4-point reduction. Mean native sequence log-confidence decreases from $-0.318$ to $-0.570$, while mean entropy increases from 0.826 to 1.382. Thus native confidence does respond to evidence loss in aggregate; the failure is not complete insensitivity.

\subsection{Native confidence is trajectory-wise non-monotonic}

\begin{figure*}[t]
    \centering
    \includegraphics[width=\textwidth]{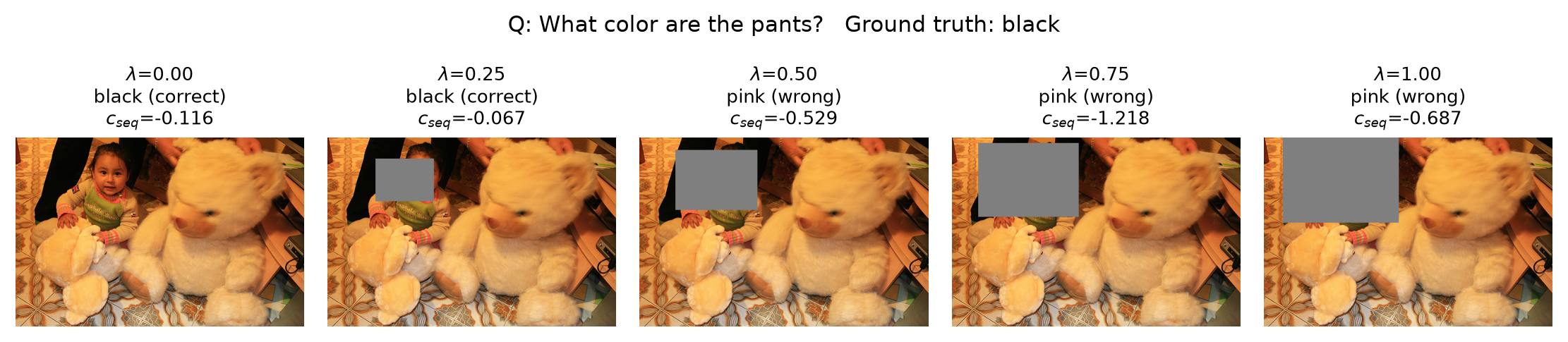}
    \caption{Example native-confidence violation. The answer is initially correct, changes after additional critical masking, and confidence is non-monotonic across the trajectory.}
    \label{fig:qualitative}
\end{figure*}

Across 704 adjacent pairs, native confidence violates the desired direction 307 times, giving EMVR 0.436. At least one violation occurs in 162/176 trajectories (92.0\%). Restricting to the 113 clean-correct trajectories lowers EMVR to 0.383, but 88.5\% still contain a violation. Figure~\ref{fig:qualitative} illustrates a non-cherry-picked failure type from the cached failure analysis: confidence first increases under light masking and later rebounds under full masking even after the answer has switched from correct to wrong.

\subsection{The degradation is evidence-specific}

\begin{figure*}[t]
    \centering
    \includegraphics[width=\textwidth]{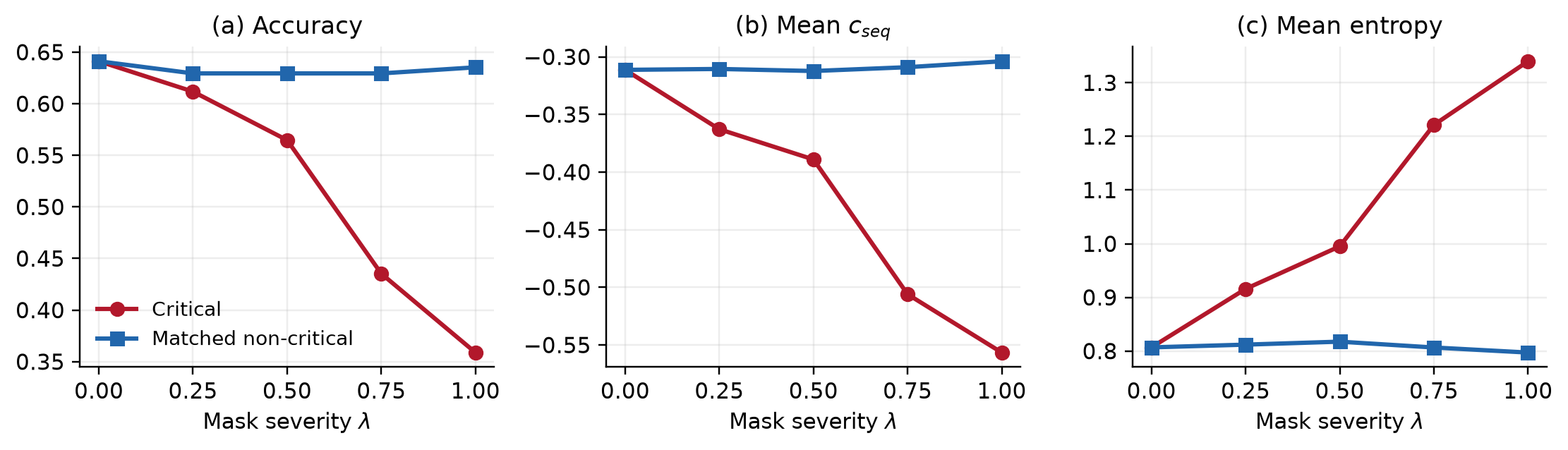}
    \caption{Critical versus matched non-critical masking on 170 paired trajectories. Non-critical curves remain nearly flat in aggregate despite a high directional EMVR caused by small fluctuations.}
    \label{fig:control}
\end{figure*}

Table~\ref{tab:control} and Figure~\ref{fig:control} compare critical and equally sized matched non-critical masks on the 170 valid pairs. Critical masking causes a 0.282 accuracy drop; non-critical masking causes 0.006. The paired accuracy-drop difference is 0.276 (95\% CI [0.200, 0.347]). Critical masks also produce substantially larger confidence decreases and entropy increases. This control supports the intended causal interpretation: the main degradation is associated with removing question-relevant evidence rather than merely covering the same number of pixels.

\begin{table}[t]
\centering
\caption{Matched control on 170 valid paired trajectories. Changes are full-mask minus clean except accuracy drop, which is clean minus full.}
\label{tab:control}
\footnotesize
\setlength{\tabcolsep}{3pt}
\begin{tabular}{lrrr}
\toprule
Intervention & Acc. clean & Acc. full & Acc. drop \\
\midrule
Critical & 0.641 & 0.359 & 0.282 \\
Matched non-critical & 0.641 & 0.635 & 0.006 \\
\midrule
Paired difference & \multicolumn{2}{r}{critical $-$ non-critical} & 0.276 \\
95\% CI & \multicolumn{2}{r}{} & [0.200, 0.347] \\
\bottomrule
\end{tabular}
\end{table}

The non-critical confidence curve is nearly flat, yet its EMVR is 0.528 versus 0.434 for critical masking. This does not make non-critical masking more harmful: when adjacent differences are small and noisy around a constant level, their directions approach chance. The primary control statistic is the paired task-degradation magnitude.

\subsection{Evidence-order supervision reduces masking violations}

Table~\ref{tab:masking} reports the frozen primary masking comparison. BCE plus order reduces EMVR from 0.330 to 0.303 and the trajectory violation rate from 0.813 to 0.767. The paired order-minus-BCE EMVR difference is $-0.026989$ (95\% CI [$-0.044034$, $-0.009943$]), supporting a structural improvement. AUROC changes by only $+0.003773$ (95\% CI [$-0.002943$, $0.010189$]) and AURC by $+0.000702$ (95\% CI [$-0.004039$, $0.007011$]); both intervals cross zero. We therefore claim reduced order violations with broadly preserved discrimination, not significant AUROC or AURC improvement.

\begin{table*}[t]
\centering
\caption{Frozen primary masking results. Lower is better for Brier, EMVR, trajectory violation rate (TVR), and AURC. Native confidence is a score, so learned-head accuracy and Brier are not applicable.}
\label{tab:masking}
\small
\begin{tabular}{lrrrrrrr}
\toprule
Score/model & Accuracy & AUROC & AUPRC & Brier & EMVR & TVR & AURC \\
\midrule
Native confidence & -- & \textbf{0.7689} & -- & -- & 0.4361 & -- & \textbf{0.2654} \\
BCE-only & \textbf{0.6104} & 0.6743 & \textbf{0.7237} & 0.2268 & 0.3298 & 0.8127 & 0.3151 \\
BCE + order & 0.6036 & 0.6762 & 0.7222 & \textbf{0.2256} & \textbf{0.3027} & \textbf{0.7673} & 0.3158 \\
\bottomrule
\end{tabular}
\end{table*}

\subsection{Ordering transfers from masking to unseen blur}

Local blur produces a 21.6-point clean-to-maximum accuracy loss (0.642 to 0.426), with native blur EMVR 0.420. In the held-out mask-train $\rightarrow$ blur-test experiment (Table~\ref{tab:transfer}), order training reduces mean EMVR from 0.449 to 0.402. The paired difference is $-0.046778$ (95\% CI [$-0.073864$, $-0.019886$]). This statistically supported transfer is the strongest evidence that the objective learns more than a mask-specific score pattern.

The AUROC difference is $+0.013338$ (95\% CI [$-0.021048$, $0.049351$]), the Brier difference is $-0.013885$ (95\% CI [$-0.043784$, $0.014721$]), and the AURC difference is $-0.013053$ (95\% CI [$-0.043227$, $0.018787$]). These conventional metric differences remain statistically inconclusive.

\begin{table*}[t]
\centering
\caption{Mask-train $\rightarrow$ blur-test grouped-cross-validation means. Lower is better for Brier, EMVR, and TVR. The statistically supported paired result is the order-minus-BCE EMVR difference; other metric intervals cross zero.}
\label{tab:transfer}
\small
\begin{tabular}{lrrrrrr}
\toprule
Score/model & Accuracy & AUROC & AUPRC & Brier & EMVR & TVR \\
\midrule
Native confidence & -- & \textbf{0.7010} & \textbf{0.7110} & -- & 0.4207 & 0.8921 \\
BCE-only & 0.5979 & 0.6049 & 0.6142 & 0.3521 & 0.4492 & 0.9089 \\
BCE + order & \textbf{0.6026} & 0.6205 & 0.6300 & \textbf{0.3383} & \textbf{0.4022} & \textbf{0.8863} \\
\midrule
Order $-$ BCE EMVR & \multicolumn{6}{c}{$-0.04678$; 95\% CI [$-0.07386$, $-0.01989$]} \\
\bottomrule
\end{tabular}
\end{table*}

\begin{figure}[t]
    \centering
    \includegraphics[width=\columnwidth]{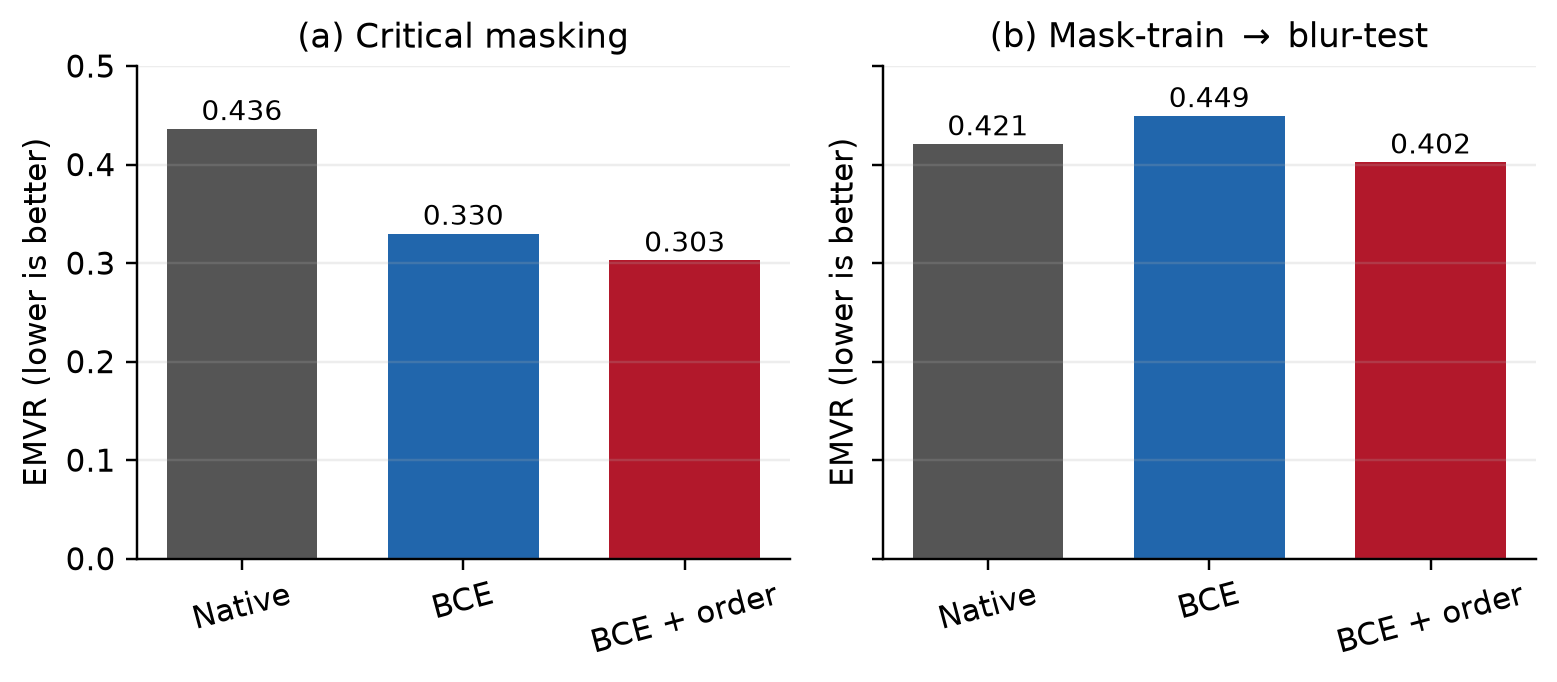}
    \caption{EMVR under the primary masking experiment and held-out blur transfer. The order head improves on BCE in both settings; native confidence is shown as a behavioral reference, not an identically trained model.}
    \label{fig:emvr}
\end{figure}

\subsection{Selective prediction: a negative result}

\begin{figure}[t]
    \centering
    \includegraphics[width=\columnwidth]{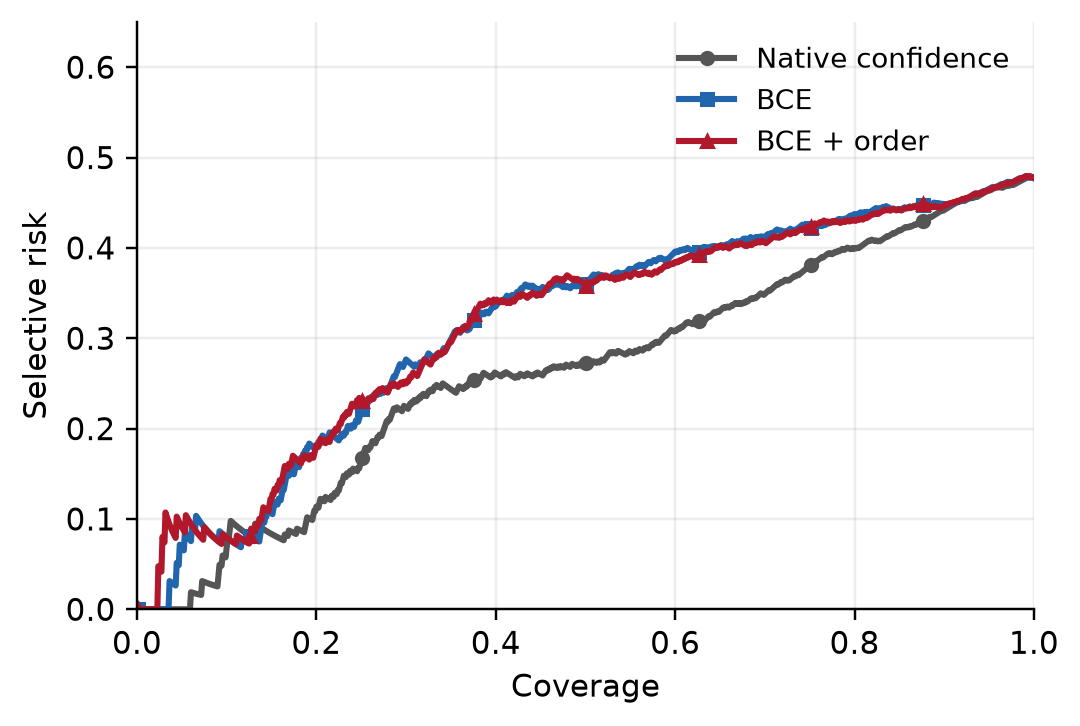}
    \caption{Masking risk--coverage curves. Native confidence ranks correctness better than either learned head despite its higher evidence-order violation rate.}
    \label{fig:risk}
\end{figure}

Native confidence remains the strongest masking selective-risk score: AURC is 0.265 versus 0.315 for BCE and 0.316 for BCE plus order. At approximately 50\% coverage, risk is 0.273 for native confidence and 0.359 for both learned heads. Figure~\ref{fig:risk} shows the complete curves. The proposed objective does not improve selective prediction over native confidence.

\subsection{Feature and objective ablations}

Table~\ref{tab:ablations} summarizes P20 ablations, which are separate from the frozen primary comparisons above. Confidence-only features discriminate correctness strongly on masking, whereas hidden-only features do not automatically become monotonic. In cross-blur transfer, hidden-only EMVR is 0.456 and full hidden-plus-signals BCE EMVR is 0.442. Adding the order objective to the same full representation yields the best structural transfer EMVR, 0.391. This supports attributing the transfer benefit to the training objective rather than assuming hidden representations are naturally ordered.

\begin{table*}[t]
\centering
\caption{P20 feature/objective ablations (means). These are diagnostic ablations, not replacements for the frozen primary results. HS denotes hidden states plus native signals.}
\label{tab:ablations}
\small
\begin{tabular}{llrr|rr}
\toprule
& & \multicolumn{2}{c}{Critical masking} & \multicolumn{2}{c}{Mask $\rightarrow$ blur} \\
Features & Objective & AUROC & EMVR & AUROC & EMVR \\
\midrule
Confidence & BCE & \textbf{0.7614} & 0.4364 & \textbf{0.7010} & 0.4207 \\
Entropy & BCE & 0.7515 & 0.4136 & 0.6829 & 0.4135 \\
Confidence + entropy & BCE & 0.7612 & 0.4278 & 0.6959 & 0.4107 \\
Hidden only & BCE & 0.6152 & 0.4093 & 0.5917 & 0.4564 \\
HS & BCE & 0.6298 & 0.4037 & 0.6068 & 0.4421 \\
HS & BCE + order & 0.6536 & \textbf{0.3950} & 0.6175 & \textbf{0.3909} \\
\bottomrule
\end{tabular}
\end{table*}

\section{Analysis and Discussion}
\label{sec:analysis}

\paragraph{Correctness discrimination is not evidence-order consistency.}
The main conceptual result is the divergence between two reliability objectives. Native confidence has masking AUROC 0.769 and the best AURC, yet its EMVR is 0.436 and 92.0\% of trajectories contain a directional violation. The order-trained head sacrifices neither metric catastrophically, but its supported improvement is specifically structural. A reliability score used to rank a mixed population of answers need not obey the within-question counterfactual order induced by evidence loss.

\paragraph{Why a flat control can have high EMVR.}
EMVR counts only the sign of each adjacent difference. For the matched non-critical control, accuracy, mean confidence, and mean entropy are almost unchanged from clean to full masking. Tiny stochastic or numerical fluctuations around a flat trajectory can nevertheless yield an upward sign about half the time. Consequently, EMVR without an effect-size curve can be misleading. The P18 conclusion rests on paired changes in accuracy, confidence, and entropy; the control EMVR is reported for completeness.

\paragraph{What transfers.}
The transfer experiment changes the intervention while retaining local question-critical geometry. The order loss trained on hard gray masking continues to reduce violations under Gaussian blur. The lack of statistically resolved AUROC, Brier, or AURC gains narrows the interpretation: the transferred object is evidence ordering, not generic uncertainty quality. The P20B ablation strengthens this reading because hidden features without the order objective transfer poorly.

\paragraph{The epistemic prior is imperfect.}
Equation~\ref{eq:desired-order} encodes a useful expectation, not a logical law of every sample. Partial masking can remove distractors, alter segmentation, or accidentally expose language priors; an answer may also remain recoverable from correlated context. We therefore do not equate every violation with a model error. EMVR is most informative in aggregate, together with task degradation and matched controls.

\paragraph{Qualitative failure modes.}
Cached trajectories exhibit several patterns: clean-correct answers can switch under degradation while confidence rebounds; some correct answers survive full masking through context or priors; wrong clean answers can remain confidently stable; matched non-critical masks often leave answers unchanged; and both blur and masking can produce non-monotonic answer switches. These cases argue against a single scalar interpretation of confidence and motivate reporting behavioral trajectories alongside aggregate discrimination metrics.

\section{Limitations}
\label{sec:limitations}

The study has several important limits. First, the frozen benchmark contains only 176 accepted trajectories from one GQA-derived subset and one primary VLM family/model. Questions are concentrated in color, identity, material, and shape categories. Evidence regions are scene-graph object boxes rather than exhaustive human rationales, and the main audit used automated visual quality control with only a limited human spot-check. Strict normalized exact matching treats semantically close responses such as ``tan'' and ``khaki'' as different; we did not add an unreported semantic-equivalence analysis.

Only local Gaussian blur is tested strongly as an unseen degradation family. The method does not improve AURC over native confidence, and learned-head AUROC/Brier differences are not statistically significant. Order constraints encode an epistemic prior that may fail for individual samples, particularly when correlated context remains. Matched non-critical boxes are equally sized and geometrically separated but can overlap other annotated objects; only 32/170 have zero annotated-object overlap. Category labels were corrected after initial processing in P17, so stale category metadata in some blur caches must not be used for category-specific claims. Finally, a larger study should test multiple VLM families, datasets, evidence annotations, and intervention mechanisms before generalizing beyond this controlled setting.

\section{Conclusion}
\label{sec:conclusion}

Progressive loss of question-critical evidence reveals a gap between aggregate confidence response and within-trajectory evidence ordering. A lightweight post-hoc head trained with correctness plus evidence-order supervision reduces that structural inconsistency under masking, and the reduction transfers statistically to unseen local blur. However, native confidence remains stronger for selective-risk ranking, while AUROC and Brier differences between learned objectives are inconclusive. Evidence-order consistency and conventional confidence quality should therefore be treated as related but distinct reliability objectives.

\bibliographystyle{plainnat}
\bibliography{references}

\appendix
\section{Additional Protocol Details}

\subsection{Benchmark audit and taxonomy}
The 250-candidate scaled set was reviewed by an automated visual-quality procedure that classified 176 examples as accepted, 71 as rejected, and 3 as unsure. A deterministic ten-example human spot-check is retained in \path{results/scaled/p17/human_spotcheck_10.json}. It must not be interpreted as full manual auditing. P17 corrected the category taxonomy: across 880 condition rows, the final counts are color 315, identity 400, shape 60, and material 105; 205 labels changed and zero correctness labels changed.

\subsection{Matched non-critical box search}
For each accepted critical box, a $41\times41$ deterministic grid of possible top-left placements is evaluated. Candidates retain exactly the critical width and height, remain in image bounds, and satisfy IoU $\le0.05$. The objective sums intersection-area/control-area ratios over all non-critical scene-graph boxes. Ties favor greater center distance from the critical region and then coordinate order. Selection uses seed 42 as experiment metadata but no randomness. Six large-box cases have no valid candidate. The valid controls have mean critical-control IoU 0.00506 and mean overlap score 0.81582; the latter can exceed one because GQA annotations are dense and nested.

\subsection{Statistics}
All stated intervals use 1,000 trajectory-level paired bootstrap resamples with seed 42. A resampled unit contains every severity state of one question under both compared methods or interventions. This preserves within-trajectory dependence and pairing. Reported intervals are percentile intervals; no condition-level independence assumption or p-value is used.

\end{document}